\documentclass[11pt,a4paper]{article}
\usepackage{times,latexsym}
\usepackage{url}
\usepackage[T1]{fontenc}

\usepackage[acceptedWithA]{tacl2020}
\usepackage[]{tacl2020}

\usepackage{times}
\usepackage{latexsym}
\usepackage{makecell}
\usepackage{graphicx}
\usepackage{multirow}
\usepackage{url}
\usepackage{enumitem}
\usepackage{verbatim}
\usepackage{caption}
\usepackage{hyperref}
\usepackage{xspace,mfirstuc,tabulary}

\newif\iftaclinstructions
\taclinstructionsfalse 
\iftaclinstructions
\renewcommand{\confidential}{}
\renewcommand{\anonsubtext}{(No author info supplied here, for consistency with
TACL-submission anonymization requirements)}
\newcommand{\instr}
\fi

\iftaclpubformat 

\else

\fi

\title{Generating Metaphoric Paraphrases}

\author{
 Kevin Stowe, Leonardo Ribeiro, Iryna Gurevych \\
 UKP Lab, Technical University of Darmstadt\\
 \hyperlink{www.ukp.tu-darmstadt.de}{www.ukp.tu-darmstadt.de} \\
\\
}

\date{}

\begin{document}
\maketitle
\begin{abstract}
    This work describes the task of \textit{metaphoric paraphrase generation}, in which we are given a literal sentence and are charged with generating a metaphoric paraphrase. We propose two different models for this task: a lexical replacement baseline and a novel sequence to sequence model, 'metaphor masking', that generates free metaphoric paraphrases. We use crowdsourcing to evaluate our results, as well as developing an automatic metric for evaluating metaphoric paraphrases. We show that while the lexical replacement baseline is capable of producing accurate paraphrases, they often lack metaphoricity, while our metaphor masking model excels in generating metaphoric sentences while performing nearly as well with regard to fluency and paraphrase quality.

\end{abstract}

\section{Introduction}

    Metaphors have long posed significant problems to researchers across a wide variety of fields. While humans seem capable of easily understanding even complex metaphors, it remains difficult to devise a formal analysis that captures the depth and breadth of meanings produced by novel metaphors. We typically think of metaphors within the Conceptual Metaphor framework \cite{lakoff-1980,lakoff-1993}, in which metaphors are based in conceptual mappings between different domains: we have cognitive concepts that can be used to represent and understand other concepts, and these mappings can be expressed linguistically to form concrete metaphoric expressions. 
    
    While there are many different computational approaches to metaphoric language, the field remains challenging and, in some areas, relatively unexplored. The variety of meanings captured by creative metaphors pose numerous problems to natural language processing researchers, as they rely on lexical diversity and conceptual knowledge. The bulk of work in metaphor has gone to identifying metaphor expressions or generating interpretations for them \cite{shutova-2015,veale-2016-2}. Whereas previous approaches focus on classification, we instead focus on generation: how can we create novel, interesting, and valid metaphoric expressions?
    
    This task has many possible applications, including creative writing assistance, where users can employ metaphor generation to develop more interesting, persuasive writing. Lakoff and Johnson \shortcite{lakoff-1980} suggest that not only can metaphors capture similarity between domains, they actually can generate the similarity, allowing us to view concepts in new ways; optimistically, metaphor generation may allow us to discover new metaphoric ideas to foster understanding and growth in scientific areas. This is particularly true in the domain of education, where new metaphors can be instructive both for teachers and students \cite{marshall-1990}. Metaphors are also critical for proper interaction between humans and computational agents: humans produce metaphors easily, and to have natural communication with computational models will require them to be able to do the same \cite{zhang-2008,wallington-2011}. 
    
    In contrast to previous work generating novel metaphors (\S \ref{sec:gen-background}), we are the first to tackle metaphor paraphrase generation, and we hope our work can function as a jumping-off point for this challenging and interesting task. This task is a particularly difficult task for a variety of reasons. First, metaphors have the potential to be enormously creative, deviating greatly from "standard" language, which means normal language models may have difficulty in producing good metaphors. Traditional paraphrasing systems attempt to keep the sentences relatively similar, while in fact we need sentences that vary substantially, in order to enforce metaphor production.
    
    This leads also to significant problems: there are countless possible metaphor paraphrases for any given utterance and there are numerous possible metaphoric mappings that can be evoked, yielding slightly different semantic connotations. Consider the following example:
    
    \begin{enumerate}
        \item The company was losing money rapidly.
    \end{enumerate}
    
    This sentence has numerable possible metaphoric paraphrases, evoking many different metaphors:
    
    \begin{enumerate}\setcounter{enumi}{1}\setlength{\itemsep}{0cm}
        \item \label{ex:1} The company was hemorrhaging money.
        \item \label{ex:2} The company's finances were circling the drain.
        \item \label{ex:3} The business fell off of a cliff.
        \item \label{ex:4} Profits collapsed.
    \end{enumerate}
    
 In \ref{ex:1} and \ref{ex:2}, "money" is conceptualized as blood and water respectively, and from conceptual metaphor theory we see that this evokes the \textsc{money is a liquid} mapping. In \ref{ex:3}, "finances" is conceptualized as a physical entity, and further, one that can experience harm, perhaps evoking the \textsc{economic harm is physical injury} mapping. In \ref{ex:4}, the company's profits are conceptualized as a building, evoking the frequent metaphor of social and economic constructs being conceptualized as physical constructions, in this case specifically \textsc{finances are buildings}.
    
 Note that there is a seemingly endless variety of metaphoric expressions that can fairly consistently capture the same general meaning, with a wide variety of lexical variation. This makes metaphoric paraphrases extremely difficult to evaluate automatically: traditional metrics for generation (such as BLEU \cite{banerjee-2005} and ROUGE \cite{papineni-2002}) rely heavily on word overlap, which is actually counterproductive for metaphoric paraphrasing: we would like our generated phrases to have less word overlap, as interesting metaphors are likely to share little lexical overlap with the original inputs. For this reason we rely on crowdsourcing, evaluating metaphoricity, fluency, and paraphrase quality.

    We approach the problem of metaphoric paraphrase generation from a variety of backgrounds, each with their own positives and negatives. First, we will consider the problem one of lexical replacement, in which we identify the important words in the literal utterance and replace them with metaphoric counterparts. This yields coherent utterances, but limits the flexibility of the output. Second, we will consider this a sequence to sequence (seq2seq) problem, and employ a novel generation technique dubbed "metaphor masking" to hide important words in the input during training and evaluation, forcing the seq2seq model to learn the appropriate contexts for metaphoric and literal words. This also requires knowledge of the key words before paraphrasing, but allows for substantially more flexibility in generation. 

    Our contribution is thus threefold:
    \begin{itemize}
    \setlength{\itemsep}{0cm}
    \item We formalize the task of metaphor generation, elucidating the datasets and experimental setup necessary.
    \item We implement a lexical replacement-based baseline, as well as a novel seq2seq architecture based on "metaphor masking".
    \item We perform analysis of generated metaphors, identifying strengths and weaknesses for each method.
    \end{itemize}
    
\section{Related Work}
    While our task is new, it bears similarity to a variety of better known NLP benchmarks. In the metaphor community, most of the efforts are focused on identification and interpretation of metaphors. We will instead focus on our two key components, paraphrasing and generation, as they relate to metaphors.
    
    \subsection{Literal Paraphrasing}
    Previous work investigates paraphrasing from metaphoric utterances to literal ones with the goal of providing interpretations \cite{mao-2018,shutova-2010}.  Shutova et al. \shortcite{shutova-2010} treats identification and interpretation jointly, and generates literal paraphrases for metaphoric adjective-noun phrases.
    Vector space models have also been employed successfully for generating literal paraphrases. Shutova et al. \shortcite{shutova-2012} identify a set of candidate paraphrases based on context and word vectors, and then use a model of selectional preferences to pick the most literal paraphrase. They require no training data, and achieve promising results for unsupervised literal paraphrasing.
    
    Similarly, Mao et al. \shortcite{mao-2018} build a metaphor identification system using word vectors, and also use it to generate paraphrases for metaphoric sentences. This is done by replacing the verbs that are identified as metaphoric with the most likely literal candidates. They use Word2Vec embeddings \cite{mikolov-2013} combined with WordNet to identify relations between literal and metaphoric lexemes. This allows for replacement of rarer, more metaphoric senses to concrete literal ones, but doesn't provide a solution for transitioning from a literal sense to an appropriate metaphoric one. Thus their work is effective at metaphoric to literal paraphrasing, but functions only in this direction; we will restructure their algorithm for the metaphoric direction as a lexical baseline in \S4.1.

    \subsection{Metaphor Generation}
    \label{sec:gen-background}
    With regard to metaphor generation, most efforts have been to generate metaphors at the lexical or phrase level, using template- and heuristic-based methods. Early work in computational metaphor generation involves generating simple "A is like B" expressions, based on probabilistic relationships between words \cite{abe-2006,terai-2010}. These methods are effective to a degree, but lack the flexibility necessary to instantiate natural language metaphors.

    Other early approaches to metaphor generation are rooted in knowledge bases. Hervas et al. \shortcite{hervas-2007} build a metaphor generation system by identifying metaphoric domains, building mappings between the source and target, and replacing appropriate references with the built metaphors. They show the difficulty of determining appropriate target domains for metaphors in context. Others use WordNet, building knowledge representations through semantic information from definitions \cite{veale-2009}.
    
        Other works seek to generate conceptual metaphors, rather than open linguistic expressions. These approaches, designed to generate conceptual metaphor mappings such as \textsc{money is a liquid}, vary from WordNet- and selectional preference-based \cite{mason-2004}, clustering over WordNet senses \cite{gandy-2013}, and using proposition databases built from syntactic relations \cite{ovchinnikova-2014}. While this task is interesting and useful, particularly for doing proper reasoning from metaphoric mappings, our goal is instead to generate natural linguistic metaphors, rather than metaphoric mappings. 

        Word embedding approaches have been popular and effective for lexical metaphor tasks. In addition to Mao et al. and Shutova et al.'s paraphrasing work, Gagliano et al. \shortcite{gagliano-2016} build off of Word2Vec, using the generated vectors to identify poetic relationships between words, developing a vector-based interpretation of conceptual blends \cite{fauconnier-1996}. They identify "connector words" between concepts, allowing for the creation of linguistic metaphors that accurately capture these conceptual metaphoric mappings.

        More recently there have been efforts using deep learning methods to generate metaphoric expressions more freely, using sequence-to-sequence models. Most notable is Yu et al. \shortcite{yu-2019}, who use neural models to generate metaphoric expressions in an unsupervised manner. They identify source and target verbs automatically from corpora, and use these to train a neural language model. Our work is similar: they encode both literal and metaphoric pairs and produce metaphoric outputs based on verbs, but their generation task is free. We are instead working on the more constrained task of generating specific paraphrases from literal utterances.
    
    This is the experimental paradigm we will be following: given a literal phrase, we generate a metaphoric paraphrase that should capture the same meaning. Unlike previous work, our methods are broadly applicable to free text: we are not limited to paraphrasing individual words or phrases, but rather use deep learning models for full natural language generation, which can then freely create literal paraphrases. To our knowledge, our work is the first to attempt to explicitly generate metaphoric paraphrases.

\section{Data}
    Our goal is to generate metaphoric paraphrases for given literal phrases. Data for this task is extremely sparse: there aren't any large scale parallel corpora containing literal and metaphoric paraphrases. Most useful is that of the Mohammad et al. \shortcite{mohammad-2016}. Their dataset includes multiple parts; importantly, it contains 171 metaphoric sentences extracted from WordNet, with manually generated literal paraphrases. These are high quality annotations, and we will use this dataset for evaluation. While originally built from the side of generating literal paraphrases for metaphoric utterances, it is easy enough to reverse the direction, using their literal paraphrases as input and attempting to generate metaphoric outputs. 
    
    Note that there are some discrepancies between the original usage and our intended paraphrase usage. Notably, the dataset was originally built around verbs: the authors replaced the key verbs in each metaphoric sentence to yield a more literal output. This ignores cases where the metaphoric meaning of the sentence is captured by components other than the verb:
    \begin{enumerate}
        \setlength{\itemsep}{0cm}
        \item The painting seems to \textbf{capture} the \textit{essence of Spring}.
        \item These events could \textbf{fracture} the \textit{balance of power}.
        \item \label{ex:moon} The new moon \textit{\textbf{reflected} back at itself} from the lake's surface.
    \end{enumerate}
    
    In these examples, the verb that was replaced to make a paraphrase is in bold, while the italic phrases could also be construed as metaphoric. In particular, \ref{ex:moon} is likely to be considered metaphoric regardless of the bolded verb, due to the poetic reflexive construction "back at itself". This means that the resulting "literal" paraphrases contain literal verbs, but the sentences themselves may still contain metaphors. This isn't prevalent in the data and doesn't impact the experiments, as we are only trying to generate more metaphoric output sentences from more literal inputs, but it is important to be aware that our paraphrasing task differs somewhat from the design of the original dataset.
    
    The size of this dataset is small: 171 instances is not enough to train viable deep learning models, and large scale parallel corpora for this task don't exist. For this reason, we will use methods that are either unsupervised, or don't rely on parallel data, and can be developed using non-parallel corpora. The lexical replacement model is the former, requiring no training data. The metaphor masking seq2seq model uses external training data, but does not require the data to be parallel. We use a masking procedure to generate artificial sentence pairs for seq2seq training, allowing the model to be function using non-parallel datasets.
    
\section{Methods}
    We propose two different models for metaphoric paraphrase generation. First, we implement a lexical replacement baseline, based on that of Mao et al. \shortcite{mao-2018}. Second, we develop a novel seq2seq framework that masks metaphoric words to better learn how to generate metaphoric outputs.
    
\subsection{Lexical Replacement Baseline}
    \label{sec:lex-replacement}

            \begin{figure}[t]
    \includegraphics[width=.5\textwidth]{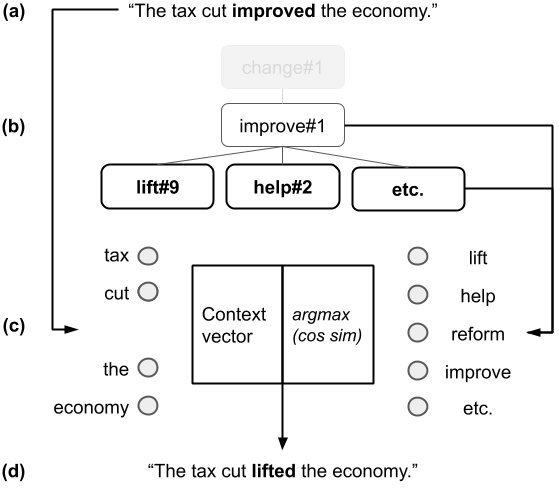}
    \centering
    \vspace{-1ex}
    \caption{\label{fig:mao} Lexical Replacement Baseline}
    \end{figure}

    Metaphors often hinge on verbs. This intuition has fueled many identification and interpretation projects, including the inclusion of the verb-specific identification track of the metaphor detection shared task \cite{leong-2018}. We implement a lexical replacement baseline that takes the literal verb and replaces it with a more metaphoric counterpart. This is based on the work of Mao et al. \shortcite{mao-2018}, who employ this strategy in the other direction: they take metaphoric sentences and replace the metaphoric verbs with literal ones. 
    
    We implement this algorithm for metaphor generation by reversing their candidate selection. For an overview of the process, see Figure \ref{fig:mao}. We begin with a literal sentence with a marked verb (a). (b) We use the WordNet sense hierarchy to find related words to the input word which will then be "candidates" to replace it, but rather than searching "up" the hierarchy for hypernyms, we search "down" the hierarchy for troponyms: more specific verbs (in bold). We believe that in the lexical replacement task, replacement with more specific verbs is likely to yield more metaphoric expressions, as these specific verbs require specific contexts to be understood literally. When placed in an unfamiliar context, they adopt metaphoric meanings via a coercion-like process \cite{steedman-1988}.  (c) We follow their algorithm for picking the best candidate: we take the mean output embedding of the context (based on the Google News Word2Vec vectors \cite{mikolov-2013}), and select the candidate word that best matches that mean by way of cosine similarity. (d) This yields the (more specific) word that best fits the context, generating a more metaphoric expression.
    
    This method, then, takes as input a sentence with a known literal verb, generates possible metaphoric candidates to replace that verb, and chooses the best fitting option. It requires no external training data, but relies on WordNet, and is restricted to only generating metaphoric verbs.

\subsection{Metaphor masking model}
        \begin{figure}[t]
    \includegraphics[width=.5\textwidth]{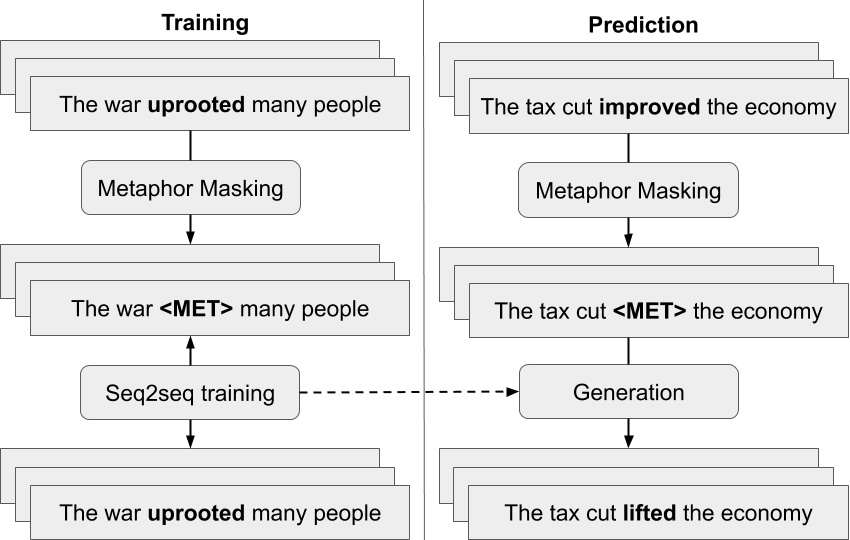}
    \centering
    \vspace{-1ex}
    \caption{\label{fig:seq2seq} Metaphor masking for the seq2seq model.}
    \end{figure}

    Sequence to sequence (seq2seq) learning paradigms are vital for a variety of NLP applications: machine translation, style transfer, natural language generation, and more \cite{chen-2018,mueller-2017,dusek-2020}. These methods rely on encoding input sentences into vectors, and then applying decoders to generate some output from that input vector. They are often trained on parallel corpora (as in the case of machine translation), with the model learning to output some text based on the vector encoded from the input.
    
    Seq2seq models have been used to generate metaphoric text \cite{yu-2019}, but here we are focused on paraphrase generation. In order to apply seq2seq models to this task, we develop a new framework dubbed "metaphor masking". In this framework, we replace metaphoric words in the input texts with metaphor masks (unique "metaphor" tokens), hiding the lexical item. This creates artificial parallel training data: the input is the masked text, with the hidden metaphorical word, and the output is the original text. Through this learning paradigm, the model learns that it needs to generate metaphoric words when it encounters the metaphor mask token. At test time, we provide the model with the literal input, mask the verb, and the model produces an output conditioned on the metaphor masking training. An overview of the process is shown in Figure \ref{fig:seq2seq}.
    
    This procedure requires additional annotated data to generate the parallel inputs for training. For this, we employ a number of available metaphor corpora: the VUAMC dataset \cite{pragglejaz-2010}, another partition of the Mohammad et al. dataset that contains individual sentences labelled as literal or metaphoric \cite{mohammad-2016}\footnote{Note that our test data also comes from this source: we have removed all text examples from this dataset for all training.}, the Trofi dataset \cite{birke-2006}, and the additional data collected by Stowe et al. \shortcite{stowe-2018}. Each of these datasets contains annotations of metaphoric verbs, although the annotation schema differ, so we expect some variety and noise in the model. Combining these datasets yields 35,415 verbs, of which 11,593 are metaphoric.

    Our final goal is to generate short metaphoric utterances based on the Mohammad et al. \shortcite{mohammad-2016} dataset. In order to match this, we trim our training data around the verbs: each verb is treated as a separate training instance, along with 7 words of context on each side.  We use all 35,415 sentences as input to the model: non-metaphoric sentences are left as-is, with the input mirroring the output. Metaphoric data is masked during training, replacing the input verb with a metaphor masking and using the original as output. This yields 35,415 pairs for training, 11,593 of which contain metaphoric masks. We hypothesize that using both literal and metaphoric datasets will allow the model to better distinguish between sentences with a metaphor mask and those without, generating stronger metaphoric outputs. We use a transformer architecture \cite{vaswani-2017} with 6 layers in the encoder and decoder. The model uses 8 heads to learn different attention distributions. In the end they are concatenated. The hidden size for encoder and decoder is 512. We use normalization per tokens, with a vocabulary size of 30K. The model was trained using ADAM optimiser, with an initial learning rate of 0.5.
    
    \section{Crowdsourced Evaluation}

    The approaches to evaluating metaphoric and literal sentences using crowdsourcing  include evaluating hand-generated sentences for metaphoricity \cite{mohammad-2016,bizzoni-2018}, evaluation of the output of automatic metaphor generation systems \cite{yu-2019,veale-2016}, and evaluation of novelty in verbal metaphors \cite{do-dinh-2018}. Uniquely focusing on metaphor evaluation, Miyazawa and Miyao \shortcite{miyazawa-2017} highlight the importance of effective evaluation. They use four key metrics: metaphoricity, novelty, comprehensibility, and overall evaluation, to measure the success of metaphor generation in Japanese. 
    
    We will rely on two components that are typical of metaphor generation. First, we evaluate metaphoricity, with the goal of producing coherent and interesting metaphors, rather than conventional, common language. Second, we evaluate fluency, attempting to capture the syntactic viability of the generated output. Additionally, as we are attempting to generate paraphrases, we also include crowdsourced evaluation of paraphrase quality.
    
    Annotators were thus asked to rate sentences with regard to three different factors: metaphoricity, fluency, and paraphrase quality. Each sentence was rated by five separate workers on a Likert scale from 1 to 4.\footnote{We chose this scale rather than the 1 to 5 used by Yu et al. \shortcite{yu-2019} to encourage workers to avoid "neutral" responses of '3'.} We filtered out results of users who failed test sentences and those who only completed 1 task, aiming to keep results from consistent and knowledgeable workers.
    
    Fluency judgments were relatively simple. For this, we asked annotators to rate the sentences based on how fluent (from incomprehensible to fluent English) a sentence is.
    
    For paraphrase judgments, we used with two different setups. We have access to three components: the original literal input $x$; $y$, the original metaphoric paraphrase of $x$; and $y'$, the generated metaphoric paraphrase of $x$. We first evaluate $y, y'$ paraphrasing, comparing the generated metaphoric outputs with the gold metaphors from the test data, allowing us to compare the system output to the gold data.  We also experimented with comparing generated paraphrases to the literal inputs, as these should also be valid paraphrases. This represents our $x, y'$ evaluation, comparing the resulting paraphrases with the original literal inputs. For each, we presented the worker with a gold input (either literal for $x, y'$ or metaphoric for $y, y'$) and the generated output, and asked them how good of a paraphrase the output was, from "completely unrelated" to "strong paraphrase".

    Metaphor evaluation is more difficult, and we attempt to follow previous crowdsourcing approaches for metaphor rating. Based on the schema from Do Dinh et al. \shortcite{do-dinh-2018} and Yu et al. \shortcite{yu-2019}, we provided basic definitions of metaphoricity for crowdworkers, allowing them to use their intuitions about what to consider metaphoric. We found in a pilot study that providing longer, more complex descriptions of metaphoricity increased the difficulty of the task, so we chose to keep the definition simple.\footnote{To facilitate future work, full descriptions of the task, parameters, payments, and guidelines, along with the crowdsourced results and codebase, will be released upon publication.}

    Our crowdsourcing setup was repeated for three outputs. The gold metaphors of Mohammad et al. \shortcite{mohammad-2016}, which also contain hand-crafted literal paraphrases, the lexical replacement baseline, and the output of our experimental system: sentences generated via seq2seq with metaphor masking. 
    
    \section{Analysis}
    \begin{figure}[t]
    \includegraphics[width=.5\textwidth]{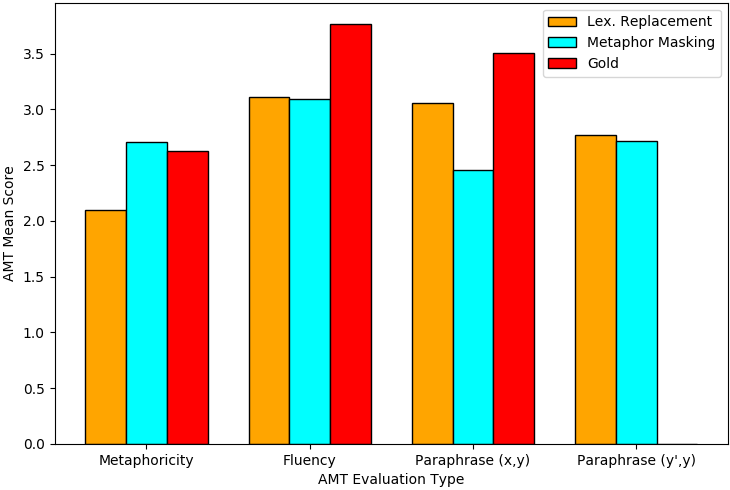}
    \centering
    \vspace{-1ex}
    \caption{\label{fig:eval} Evaluation of each model via crowdsourcing. 
    }
    \end{figure}
    
    \begin{table*}[tb]
\begin{center}
\begin{tabular} {|c|c|c|c|c|c|}
    \hline
     & \textbf{Source} & \textbf{Text} & \textbf{Met} & \textbf{Flu} & \textbf{PP} \\
    \hline
    \hline

    \multirow{4}{*}{1} & Input & He was \textbf{lavished} with praise & \multicolumn{3}{c|}{} \\\cline{2-6}
    & Gold & He was showered with praise & 3 & 4 & 4 \\
    & LexRep & He was lavished with praise & 2.6 & 3.8 & 3.8 \\
    & MM & He was pleading with impishly & 2.8 & 2.2 & 1 \\
    \hline
    \hline
    \multirow{4}{*}{2} & Input & The moon \textbf{reflected} back at itself from the lake's surface & \multicolumn{3}{c|}{} \\\cline{2-6}
     & Gold & The moon glared back at itself from the lake's surface & 3.3 & 4 & 3 \\
    & LexRep & The moon sparkled back at itself from the lake's surface & 3.2 & 4 & 3.6 \\
    & MM & The DMZ falls back at itself from the glittering surface & 3.25 & 2.6 & 1.4 \\
    \hline
    \hline
    \multirow{4}{*}{3} & Input & She \textbf{appears} among royalty & \multicolumn{3}{c|}{} \\\cline{2-6}
    & Gold & She circulates among royalty & 2.4 & 4 & 3.6 \\
    & LexRep & She manifested among royalty & 2.75 & 3.8 & 3.4 \\
    & MM & She clawed among doorknob & 2.8 & 2.6 & 1.6 \\
    \hline
    \hline
    \multirow{4}{*}{4} & Input & Life in the camp \textbf{weakened} him & \multicolumn{3}{c|}{} \\\cline{2-6}
    & Gold & Life in the camp drained him & 2.75 & 3.75 & 3.6 \\
    & LexRep & Life in the camp emasculated him & 3.3 & 3.8 & 2.8 \\
    & MM & Life in the camp miss him & 1.8 & 2.2 & 3 \\
    \hline
    \end{tabular}
\end{center}
\caption{\label{tab:lr-analysis}Samples with the highest scores for LexRep}
\end{table*}

    The mean scores for the crowdsourced evaluations for each system are shown in Figure \ref{fig:eval}.\footnote{Note that $y, y'$ comparison is unavailable for the gold data: there is no $y'$ to compare to} The sentences generated by lexical replacement bear closer resemblance to the literal inputs: they have lower metaphoricity scores and higher $x, y'$ paraphrase rankings. This is expected, as the change to the input is only a single word. The metaphor masking model shows more similarity to the metaphoric outputs: they have low paraphrase similarity in the $x, y'$ settings, but better $y, y'$ paraphrase scores, and high metaphoricity. The fact that the metaphor masking model produces metaphoricity scores on par with the gold standard metaphors, and is consistent with the lexical model in terms of fluency and $y, y'$ paraphrase quality, shows that this method is very effective at generating metaphoric sentences. The quality of the paraphrases is still relatively low, averaged 1.3 points below the gold $x, y$ paraphrases, but this is an important first step in for this task.
    
    In order to understand how each of these models performed, we do qualitative analysis over the results. We examined the results of each model: what does the lexical replacement baseline do well, and what benefits can we gain from employing our metaphor masking model?
    
    \subsection{Lexical Replacement}
    
      \begin{table*}[tb]
           \begin{center}
\begin{tabular} {|c|c|c|c|c|c|}
 \hline
     & \textbf{Source} & \textbf{Text} & \textbf{Met} & \textbf{Flu} & \textbf{PP} \\
    \hline
    \hline
    \multirow{4}{*}{5} & Input & She was \textbf{saddened} by his refusal of her invitation & \multicolumn{3}{c|}{} \\\cline{2-6}
    & Gold & She was crushed by his refusal of her invitation & 2.4 & 4 & 4 \\
    & LexRep & She was saddens by his refusal of her invitation & 1.25 & 2.8 & 3.25 \\
    & MM & She besieged by his refusal of her invitation & 3.2 & 3.8 & 3.6 \\
    \hline
    \hline
    \multirow{4}{*}{6} & Input & The critics \textbf{overpraised} this broadway production & \multicolumn{3}{c|}{} \\\cline{2-6}
    & Gold & The critics puffed up this broadway production & 3.2 & 3.75 & 3.75 \\
    & LexRep & The critics this broadway production & 2.2 & 2.25 & 2.2 \\
    & MM & The critics hailed this airborne production & 3.2 & 3.75 & 3 \\
    \hline
    \hline
    \multirow{4}{*}{7} & Input & The company \textbf{dismissed} him after many years of service & \multicolumn{3}{c|}{} \\\cline{2-6}
    & Gold & The company dumped him after many years of service & 2.2 & 4 & 3.75 \\
    & LexRep & The company scoff him after many years of service & 1.5 & 3 & 2.4 \\
    & MM & The company downsized him after many years of service & 2.6 & 4 & 3.5 \\
    \hline
    \hline
    \multirow{4}{*}{8} & Input & This story will \textbf{intrigue} you & \multicolumn{3}{c|}{}\\\cline{2-6}
    & Gold & This story will grab you & 3.2 & 3.75 & 4 \\
    & LexRep & This story will schemed you & 2.8 & 2.33 & 1.8 \\
    & MM & This story will help you & 3.6 & 4 & 2.4 \\
    \hline

    \end{tabular}
\end{center}
\caption{\label{tab:mm-analysis} Samples with the highest scores for Metaphor Masking}
\end{table*}

Table \ref{tab:lr-analysis} shows the best lexical replacement outputs, based on their improvement over the metaphor masking model. The replacement model performs well in fluency and paraphrase quality, particularly because it copies most of the input, only replacing a single word. In some cases, the "best fit" candidate is the original input word. These perform exceedingly well in fluency and paraphrase quality, as they match the input sentence, but understandably lack metaphoricity (1). However, in many cases the model often makes novel and metaphoric word choices, indicating the validity of this approach for metaphor generation (2-4).

        This baseline has numerous theoretical advantages and disadvantages. It yields output sentences that are very similar to the inputs, as we are only replacing a single word. This can be beneficial, as the outputs will be necessarily syntactically and semantically coherent except for the replaced word, but also severely restricts the creativity and novelty of the output.
    
    A downside is that this method requires knowledge of the target verb. Our data has the target verb in the literal and metaphoric paraphrases annotated, but sometimes these verbs contain particles (such as "start on" and "use up"), which make lexical replacement difficult. Just replacing the verb and maintaining the particle sometimes yields good results ("I [started] on the problem" $\rightarrow$ "I [fell] on the problem"), while replacing both verb and particle can also be correct ("they [used up] their food" $\rightarrow$ "they [demolished] their food"). Second, if we apply this method to unseen data, we will first need to identify the target verbs, making it more reliant on external knowledge and prone to error. Finally, it is dependent on WordNet, which restricts the power and flexibility with regard to creativity.
    
    While the above examples highlight the strength of the lexical replacement baseline, they also show the weaknesses of the metaphor masking approach. Due to the free nature of generation, we often see words in the generated output that bear little to relation to the original input ("impishly" in (1) and "DMZ" in (2)). These kinds of errors elucidate how the more constrained lexical replacement model tends to yield better paraphrases.

\subsection{Metaphor Masking}

The metaphor masking model tend to generate more metaphoric sentences with similar fluency, although they often are not valid paraphrases of the original input. Table \ref{tab:mm-analysis} shows examples for which the metaphor masking model performs best in comparison with the lexical replacement model.

               \begin{table*}[tb]
\begin{center}
\begin{tabular} {|c|c|c|c|c|c|}

    \hline
     & \textbf{Source} & \textbf{Text} & \textbf{Met} & \textbf{Flu} & \textbf{PP} \\
    \hline
    \hline
    \multirow{4}{*}{9} & Input & I can't \textbf{cope} with it anymore & \multicolumn{3}{c|}{} \\\cline{2-6}
    & Gold & I can't hack it anymore & 1.5 & 3 & 3.75 \\
    & LexRep & I can't improvising with it anymore & 1.8 & 2.3 & 2 \\
    & MM & I chemotherapy stuck with it gathers & 2.25 & 2 & 1.5 \\
    \hline
    \hline
    \multirow{4}{*}{10} & Input & Actions \textbf{communicate} louder than words & \multicolumn{3}{c|}{} \\\cline{2-6}
    & Gold & Actions talk louder than words & 3.2 & 3.5 & 3.75 \\
    & LexRep & Actions conveys louder than words & 1.75 & 3.4 & 3 \\
    & MM & Seurat culminated pointillism than words & 2.5 & 2 & 1.2 \\
    \hline
    \hline
    \multirow{4}{*}{11} & Input & Which horse are you \textbf{betting} on & \multicolumn{3}{c|}{} \\\cline{2-6}
    & Gold & Which horse are you backing & 2 & 4 & 4 \\
    & LexRep & Which horse are you bet on & 1.4 & 2.6 & 3.8 \\
    & MM & Gillette horse are you going on & 3 & 1.4 & 1.6 \\
    \hline
    \hline
    \multirow{4}{*}{12} & Input & A weather vane \textbf{tops} the building &  \multicolumn{3}{c|}{} \\\cline{2-6}
    & Gold & A weather vane crowns the building & 1.75 & 3.8 & 3.8 \\
    & LexRep & A weather vane clears the building & 2.25 & 4 & 2.4 \\
    & MM & A weather verbs raided the building & 3.4 & 1.8 & 1.2 \\
    \hline
    \hline
    \multirow{4}{*}{13} & Input & It \textbf{occurred} to him that she had betrayed him &  \multicolumn{3}{c|}{} \\\cline{2-6}
    & Gold & It dawned on him that she had betrayed him & 2.2 & 4 & 3.7 \\
    & LexRep & It intervened to him that she had betrayed him & 2 & 3 & 2.8 \\
    & MM & It comes to him that she had greeted him & 2.2 & 2.8 & 1.5 \\
    \hline
\end{tabular}
\end{center}
\caption{\label{tab:bad-analysis}Sentences for which both LexRep and MM models performed poorly.}
\end{table*}

Metaphor masking tends to produce fairly consistent outputs, which are syntactically regular. Hiding the metaphoric word causes the model to make a prediction, yielding varied outputs, and these are more metaphoric than their inputs. 

This model is complementary to the lexical replacement model: as it is based on a sequence-to-sequence transformer model, it is relatively free in its generation. It frequently generates words not in the original input which leads to more creative, metaphoric outputs. Examples like 5 and 6 show the power of the metaphor masking model: it is capable of generating a wider variety of words that yield better metaphors. As the model isn't constrained to a particular resource, it has more power with regard to lexical choice. Example 6 shows another benefit with regard to metaphoricity: the model can generate multiple words not present in the input ("hailed", "airborne"), yielding more creative utterances, although these are often worse paraphrases.\footnote{Note that the lexical replacement model fails for this sentence, as no candidate word was found for "puffed up".}.

While the model often generates strong metaphors, there are also cases where the model predicts a word for the masked metaphoric word that is extremely literal (7 and 8), which yields sentences that are fluent and good paraphrases but lacking in metaphoricity. This deficit is due to the lack of information for the model about the metaphoric class. As our dataset is limited, the model doesn't have enough signal to fully distinguish what goes into a metaphoric gap. More data (both metaphoric and overtly literal) should help the model generate more surprising and metaphoric outputs.

    We can also see from Table \ref{tab:mm-analysis} the weaknesses of the lexical replacement baseline. As the candidates are generated with diverse syntactic endings, they often exhibit disagreement with their arguments (5, 7, 8). Additionally, it doesn't always make metaphoric predictions: in 5, the output matches the input verb, yielding an extremely literal paraphrase.

\subsection{Consistent Errors}

The sentences that confound both of our models tend to be idiomatic (Table \ref{tab:bad-analysis}). These are cases where the "metaphoric" meaning of the sentence isn't captured explicitly by the verb, but rather spans the entire utterance. For example, in 10, the communication metaphors is present, regardless of the verb used: the literal verb "communicate" may be less metaphoric as a verb than the gold "talk", but the metaphor of the sentence persists. This causes difficulties for our systems which require metaphoricity to be focused on the verb.

The lexical replacement model often makes lexical choices that either don't match the original meaning (9), or don't maintain any metaphoricity (13). As WordNet is a finite resource, the number of candidate replacement verbs is often small, and this restricts the system from finding truly novel metaphoric expressions. It also may be the case that finding the "best fit" word from output vectors is actually counterproductive: Mao et al. \shortcite{mao-2018} use this procedure for finding the best literal paraphrases, and although we alter their approach to identify more metaphoric candidates, the model might still prefer the most literal option.

A possible solution left for future work is to select the "worst fit" from the candidates: the word who's vector is least likely to match the context. This would ensure contrast between domains, but in preliminary studies lead to the model invariably picking syntactically incomprehensible or semantically incoherent choices. For future work, we believe better limitations on the candidate selection, enforcing syntactic constraints while allowing a wider variety of domains, will allow us to implement the "worst fit" approach more effectively with the potential to generate much more interesting metaphoric replacements.

The metaphoric masking model struggles with short sentences: it often generates words that don't fit the context, yielding unparseable expressions (see 9-12). The relatively idiomatic nature of these expressions also hinders the model's performance: as the metaphoricity isn't focused singularly on the verb, the model is unable to make accurate predictions about the masked token. 

A possible solution here is to expand the masking to other parts of speech, or even to phrases. This would allow the model to generate over more complex metaphoric expressions. Additionally, if our seq2seq model can accurately pick up on masked metaphor tasks, this gives us both flexibility and control over metaphor generation: we will be able to choose which parts of utterances we would like to metaphoric, allowing for much more powerful generation systems.

    One consistent problem in this process is the difficulty of keeping annotation categories independent. We find that generated sentences that are incoherent syntactically also tend to be considered bad paraphrases (Spearman correlation of .559, $p<.01$). It is likely because if a sentence is difficult to syntactically parse, it is more difficult to assess its meaning, making judgments of semantic similarity difficult. Additionally, metaphoricity ratings correlate negatively to a lesser degree with paraphrase quality (-.112, $p\approx.03$). Strong metaphoric paraphrases likely add additional meaning or de-emphasize some of the original literal meaning, making their paraphrase quality lower. Interestingly, fluency and metaphor ratings did not significantly correlate, indicating that disfluent sentences were neither more or less metaphoric than their fluent counterparts.
    
    It is important to note the variety of possible generated expressions that are considered good. Different generated metaphors can even maintain some of the original literal meaning, while highlighting different aspects, as good novel metaphors are known for. Consider the generated example "This idea \textit{harmonizes} up with the other one", intended to paraphrase "This idea \textit{matches} up with the other one". This captures in many senses the original input of "matches up", but also provides something more: not only do the ideas go together, but perhaps they also improve upon one another. Because of the variety of acceptable outputs, automatic generation of metaphoric paraphrases is exceedingly difficult. For this reason, we present an automatic metric for evaluating metaphoric paraphrases.

\section{Conclusions and Future Work}
    We've established a new task for natural language generation: the creation of metaphoric paraphrases for literal sentences. We explore two possible models for accomplishing this task: an adapted lexical replacement baseline model that relies on WordNet to find candidate verbs and the output vectors of word embeddings to match their contexts, and a seq2seq transformer-based model that masks metaphoric verbs to encourage generation of metaphoric outputs. Crowdsourced evaluations show that both models are successful at different aspects of the task: the lexical replacement baseline yields consistent paraphrases that lack metaphoricity, while the metaphor masking model yields extremely metaphor outputs that often don't accurately paraphrase the input. 
    
    Future work in this area is hindered by the lack of available data. In order to improve these methods, we need better datasets. This couples with the problem of evaluation: standard evaluation metrics for language generation are often misleading with regard to metaphors. Better datasets would allow for the development of better metrics for evaluation, and in turn better evaluation metrics may allow us to build better systems for automatically identifying metaphoric paraphrases, allowing us to build better corpora.

    Another possible direction to explore is the incorporation of knowledge representations. Our lexical replacement method relies heavily on WordNet, and can make local changes based on a small number of candidate verbs. Our metaphor masking model is relatively free, but neither contain any knowledge of the metaphors in use. 
    
    To truly be able to generate metaphors based on actual metaphoric mappings, we need to incorporate some knowledge of the source and target domains involved. This could involve leveraging FrameNet \cite{baker-1998} or MetaNet \cite{dodge-2015}, developing a novel metaphor knowledge base, or learning domain knowledge in an unsupervised fashion. Developing metaphor knowledge bases that capture relations between domains in a usable way will not only allow for better metaphor generation, but also better reasoning and understanding of texts that make use of more complicated metaphoric expressions. However the ordeal is undertaken, generation of coherent metaphors will inevitably require better representation of the interaction between the domains evoked.

\bibliography{tacl2020}

\begin{thebibliography}{39}
\expandafter\ifx\csname natexlab\endcsname\relax\def\natexlab#1{#1}\fi

\bibitem[{Abe et~al.(2006)Abe, Kayo, and Nakagawa}]{abe-2006}
Keiga Abe, Sakamoto Kayo, and Masanori Nakagawa. 2006.
\newblock A computational model of the metaphor generation process.
\newblock In \emph{Proceedings of the 28th Annual Meeting of the Cognitive
  Science Society}, pages 937--942.

\bibitem[{Baker et~al.(1998)Baker, Fillmore, and Lowe}]{baker-1998}
C.~F. Baker, C.J. Fillmore, and J.B. Lowe. 1998.
\newblock \href {https://dl.acm.org/doi/10.3115/980845.980860} {{The Berkeley
  FrameNet project}}.
\newblock In \emph{Proceedings of the 36th Annual Meeting of the Association
  for Computational Linguistics and the 17th International Conference on
  Computational Linguistics}, pages 86--90, Montreal, Canada. Association for
  Computational Linguistics.

\bibitem[{Banerjee and Lavie(2005)}]{banerjee-2005}
Satanjeev Banerjee and Alon Lavie. 2005.
\newblock \href {https://www.aclweb.org/anthology/W05-0909} {{METEOR}: An
  automatic metric for {MT} evaluation with improved correlation with human
  judgments}.
\newblock In \emph{Proceedings of the {ACL} Workshop on Intrinsic and Extrinsic
  Evaluation Measures for Machine Translation and/or Summarization}, pages
  65--72, Ann Arbor, Michigan. Association for Computational Linguistics.

\bibitem[{Birke and Sarkar(2006)}]{birke-2006}
Julia Birke and Anoop Sarkar. 2006.
\newblock \href {https://www.aclweb.org/anthology/E06-1042} {A clustering
  approach for nearly unsupervised recognition of nonliteral language}.
\newblock In \emph{Proceedings of the 11th Conference of the {E}uropean Chapter
  of the Association for Computational Linguistics}, pages 329--336, Trento,
  Italy. Association for Computational Linguistics.

\bibitem[{Bizzoni and Lappin(2018)}]{bizzoni-2018}
Yuri Bizzoni and Shalom Lappin. 2018.
\newblock \href {https://doi.org/10.18653/v1/W18-0906} {Predicting human
  metaphor paraphrase judgments with deep neural networks}.
\newblock In \emph{Proceedings of the Workshop on Figurative Language
  Processing}, pages 45--55, New Orleans, Louisiana. Association for
  Computational Linguistics.

\bibitem[{Chen et~al.(2018)Chen, Firat, Bapna, Johnson, Macherey, Foster,
  Jones, Parmar, Schuster, Chen, Wu, and Hughes}]{chen-2018}
Mia~Xu Chen, Orhan Firat, Ankur Bapna, Melvin Johnson, Wolfgang Macherey,
  George Foster, Llion Jones, Niki Parmar, Mike Schuster, Zhifeng Chen, Yonghui
  Wu, and Macduff Hughes. 2018.
\newblock \href {http://arxiv.org/abs/1804.09849} {The best of both worlds:
  Combining recent advances in neural machine translation}.
\newblock cs.CL/1804.09849v2.

\bibitem[{Do~Dinh et~al.(2018)Do~Dinh, Wieland, and Gurevych}]{do-dinh-2018}
Erik-L{\^a}n Do~Dinh, Hannah Wieland, and Iryna Gurevych. 2018.
\newblock \href {https://doi.org/10.18653/v1/D18-1171} {Weeding out
  conventionalized metaphors: A corpus of novel metaphor annotations}.
\newblock In \emph{Proceedings of the 2018 Conference on Empirical Methods in
  Natural Language Processing}, pages 1412--1424, Brussels, Belgium.
  Association for Computational Linguistics.

\bibitem[{Dodge et~al.(2015)Dodge, Hong, and Stickles}]{dodge-2015}
Ellen Dodge, Jisup Hong, and Elise Stickles. 2015.
\newblock \href {http://www.aclweb.org/anthology/W15-1405} {Metanet: Deep
  semantic automatic metaphor analysis}.
\newblock In \emph{Proceedings of the Third Workshop on Metaphor in NLP}, pages
  40--49, Denver, Colorado. Association for Computational Linguistics.

\bibitem[{Du\v{s}ek et~al.(2020)Du\v{s}ek, Novikova, and Rieser}]{dusek-2020}
Ond\v{r}ej Du\v{s}ek, Jekaterina Novikova, and Verena Rieser. 2020.
\newblock \href {https://doi.org/https://doi.org/10.1016/j.csl.2019.06.009}
  {Evaluating the state-of-the-art of end-to-end natural language generation:
  The {E2E} {NLG} challenge}.
\newblock \emph{Computer Speech \& Language}, 59:123--156.

\bibitem[{Fauconnier and Turner(1996)}]{fauconnier-1996}
Gilles Fauconnier and Mark Turner. 1996.
\newblock Blending as a central process of grammar.
\newblock In Adele Goldberg, editor, \emph{Conceptual Structure, Discourse, and
  Language}. Cambridge University Press.

\bibitem[{Gagliano et~al.(2016)Gagliano, Paul, Booten, and
  Hearst}]{gagliano-2016}
Andrea Gagliano, Emily Paul, Kyle Booten, and Marti~A. Hearst. 2016.
\newblock \href {https://doi.org/10.18653/v1/W16-0203} {Intersecting word
  vectors to take figurative language to new heights}.
\newblock In \emph{Proceedings of the Fifth Workshop on Computational
  Linguistics for Literature}, pages 20--31, San Diego, California, USA.
  Association for Computational Linguistics.

\bibitem[{Gandy et~al.(2013)Gandy, Allan, Atallah, Frieder, Howard, Kanareykin,
  Koppel, Last, Neuman, and Argamon}]{gandy-2013}
Lisa Gandy, Nadji Allan, Mark Atallah, Ophir Frieder, Newton Howard, Sergey
  Kanareykin, Moshe Koppel, Mark Last, Yair Neuman, and Shlomo Argamon. 2013.
\newblock \href
  {https://www.aaai.org/ocs/index.php/AAAI/AAAI13/paper/view/6398} {Automatic
  identification of conceptual metaphors with limited knowledge}.
\newblock In \emph{Proceedings of the 27th AAAI Conference on Artificial
  Intelligence}, pages 328--334, Bellevue, Washington. AAAI Press.

\bibitem[{Herv\'{a}s et~al.(2007)Herv\'{a}s, Costa, Costa, Gerv\'{a}s, and
  Pereira}]{hervas-2007}
Raquel Herv\'{a}s, Rui~P. Costa, Hugo Costa, Pablo Gerv\'{a}s, and Francisco~C.
  Pereira. 2007.
\newblock Enrichment of automatically generated texts using metaphor.
\newblock In \emph{Proceedings of the Sixth Mexican International Conference on
  Artificial Intelligence}, pages 944--954, Aguascalientes, Mexico. Springer.

\bibitem[{Lakoff(1993)}]{lakoff-1993}
George Lakoff. 1993.
\newblock The contemporary theory of metaphor.
\newblock In Andrew Ortony, editor, \emph{Metaphor and Thought}, pages
  202--251. University Press Cambridge.

\bibitem[{Lakoff and Johnson(1980)}]{lakoff-1980}
George Lakoff and Mark Johnson. 1980.
\newblock \emph{Metaphors We Live By}.
\newblock University of Chicago Press, Chicago.

\bibitem[{Leong et~al.(2018)Leong, Beigman~Klebanov, and Shutova}]{leong-2018}
Chee Wee~(Ben) Leong, Beata Beigman~Klebanov, and Ekaterina Shutova. 2018.
\newblock \href {https://doi.org/10.18653/v1/W18-0907} {A report on the 2018
  {VUA} metaphor detection shared task}.
\newblock In \emph{Proceedings of the Workshop on Figurative Language
  Processing}, pages 56--66, New Orleans, Louisiana. Association for
  Computational Linguistics.

\bibitem[{Mao et~al.(2018)Mao, Lin, and Guerin}]{mao-2018}
Rui Mao, Chenghua Lin, and Frank Guerin. 2018.
\newblock \href {https://doi.org/10.18653/v1/P18-1113} {Word embedding and
  {W}ord{N}et based metaphor identification and interpretation}.
\newblock In \emph{Proceedings of the 56th Annual Meeting of the Association
  for Computational Linguistics}, pages 1222--1231, Melbourne, Australia.
  Association for Computational Linguistics.

\bibitem[{Marshall(1990)}]{marshall-1990}
Hermine~H. Marshall. 1990.
\newblock \href {https://doi.org/10.1080/00405849009543434} {This issue:
  Metaphors we learn by}.
\newblock \emph{Theory Into Practice}, 29(2):70--70.

\bibitem[{Mason(2004)}]{mason-2004}
Zachary~J. Mason. 2004.
\newblock \href {https://doi.org/10.1162/089120104773633376} {{C}or{M}et: A
  computational, corpus-based conventional metaphor extraction system}.
\newblock \emph{Computational Linguistics}, 30(1):23--44.

\bibitem[{Mikolov et~al.(2013)Mikolov, Chen, Corrado, and Dean}]{mikolov-2013}
Tomas Mikolov, Kai Chen, Greg Corrado, and Jeffrey Dean. 2013.
\newblock \href
  {http://dblp.uni-trier.de/db/journals/corr/corr1301.html#abs-1301-3781}
  {Efficient estimation of word representations in vector space}.
\newblock \emph{CoRR}, abs/1301.3781.

\bibitem[{Miyazawa and Miyao(2017)}]{miyazawa-2017}
Akira Miyazawa and Yusuke Miyao. 2017.
\newblock \href {https://www.aclweb.org/anthology/W17-6929} {Evaluation metrics
  for automatically generated metaphorical expressions}.
\newblock In \emph{The 12th International Conference on Computational
  Semantics}, Montpellier, France. Association for Computational Linguistics.

\bibitem[{Mohammad et~al.(2016)Mohammad, Shutova, and Turney}]{mohammad-2016}
Saif Mohammad, Ekaterina Shutova, and Peter Turney. 2016.
\newblock \href {https://doi.org/10.18653/v1/S16-2003} {Metaphor as a medium
  for emotion: An empirical study}.
\newblock In \emph{Proceedings of the Fifth Joint Conference on Lexical and
  Computational Semantics}, pages 23--33, Berlin, Germany. Association for
  Computational Linguistics.

\bibitem[{Mueller et~al.(2017)Mueller, Gifford, and Jaakkola}]{mueller-2017}
Jonas Mueller, David Gifford, and Tommi Jaakkola. 2017.
\newblock \href {http://proceedings.mlr.press/v70/mueller17a.html} {Sequence to
  better sequence: Continuous revision of combinatorial structures}.
\newblock In \emph{Proceedings of the 34th International Conference on Machine
  Learning}, pages 2536--2544, Sydney, Australia. PMLR.

\bibitem[{Ovchinnikova et~al.(2014)Ovchinnikova, Zaytsev, Wertheim, and
  Israel}]{ovchinnikova-2014}
Ekatarina Ovchinnikova, Vladimir Zaytsev, Suzanne Wertheim, and Ross Israel.
  2014.
\newblock \href {https://arxiv.org/abs/1409.7619} {Generating conceptual
  metaphors from proposition stores}.
\newblock cs.CL/1409.7619.

\bibitem[{Papineni et~al.(2002)Papineni, Roukos, Ward, and Zhu}]{papineni-2002}
Kishore Papineni, Salim Roukos, Todd Ward, and Wei-Jing Zhu. 2002.
\newblock \href {https://doi.org/10.3115/1073083.1073135} {{B}leu: a method for
  automatic evaluation of machine translation}.
\newblock In \emph{Proceedings of the 40th Annual Meeting of the Association
  for Computational Linguistics}, pages 311--318, Philadelphia, Pennsylvania.
  Association for Computational Linguistics.

\bibitem[{Shutova(2010)}]{shutova-2010}
Ekaterina Shutova. 2010.
\newblock \href {http://www.aclweb.org/anthology/N10-1147} {{Automatic Metaphor
  Interpretation as a Paraphrasing Task}}.
\newblock In \emph{The 2010 Annual Conference of the North American Chapter of
  the Association for Computational Linguistics}, pages 1029--1037, Los
  Angeles, California. Association for Computational Linguistics.

\bibitem[{Shutova(2015)}]{shutova-2015}
Ekaterina Shutova. 2015.
\newblock {Design and Evaluation of Metaphor Processing Systems}.
\newblock \emph{Computational Linguistics}, 41:579--623.

\bibitem[{Shutova et~al.(2012)Shutova, Van~de Cruys, and
  Korhonen}]{shutova-2012}
Ekaterina Shutova, Tim Van~de Cruys, and Anna Korhonen. 2012.
\newblock \href {https://www.aclweb.org/anthology/C12-2109} {Unsupervised
  metaphor paraphrasing using a vector space model}.
\newblock In \emph{Proceedings of the 24th International Conference on
  Computational Linguistics}, pages 1121--1130, Mumbai, India. COLING 2012
  Organizing Committee.

\bibitem[{Steedman and Moens(1988)}]{steedman-1988}
Mark Steedman and Marc Moens. 1988.
\newblock \href {https://www.aclweb.org/anthology/J88-2003} {Temporal ontology
  and temporal reference}.
\newblock \emph{Computational Linguistics}, 2(14):15--28.

\bibitem[{Steen et~al.(2010)Steen, Dorst, Herrmann, Kaal, Krennmayr, and
  Pasma}]{pragglejaz-2010}
G.J. Steen, A.G. Dorst, J.B. Herrmann, A.A. Kaal, T.~Krennmayr, and T.~Pasma.
  2010.
\newblock \emph{A method for linguistic metaphor identification. From MIP to
  MIPVU.}
\newblock Converging Evidence in Language and Communication Research. John
  Benjamins.

\bibitem[{Stowe and Palmer(2018)}]{stowe-2018}
Kevin Stowe and Martha Palmer. 2018.
\newblock \href {https://www.aclweb.org/anthology/W18-0903.pdf} {Leveraging
  syntactic constructions for metaphor processing}.
\newblock In \emph{Workshop on Figurative Language Processing}, pages 17--26,
  New Orleans, Louisiana. Association for Computational Linguistics.

\bibitem[{Terai and Nakagawa(2010)}]{terai-2010}
Asuka Terai and Masanori Nakagawa. 2010.
\newblock \href {https://rdcu.be/b2iMr} {A computational system of metaphor
  generation with evaluation mechanism}.
\newblock In \emph{International Conference on Artificial Neural Networks},
  pages 142--147, Thessaloniki, Greece. Springer.

\bibitem[{Vaswani et~al.(2017)Vaswani, Shazeer, Parmar, Uszkoreit, Jones,
  Gomez, Kaiser, and Polosukhin}]{vaswani-2017}
Ashish Vaswani, Noam Shazeer, Niki Parmar, Jakob Uszkoreit, Llion Jones,
  Aidan~N. Gomez, {\L}ukasz Kaiser, and Illia Polosukhin. 2017.
\newblock \href
  {https://papers.nips.cc/paper/7181-attention-is-all-you-need.pdf} {Attention
  is all you need}.
\newblock In \emph{31st Conference on Neural Information Processing Systems},
  pages 5998--6008, Long Beach, California. Curran Associates, Inc.

\bibitem[{Veale(2016)}]{veale-2016}
Tony Veale. 2016.
\newblock \href {https://doi.org/10.18653/v1/W16-1105} {Round up the usual
  suspects: Knowledge-based metaphor generation}.
\newblock In \emph{Proceedings of the Fourth Workshop on Metaphor in {NLP}},
  pages 34--41, San Diego, California. Association for Computational
  Linguistics.

\bibitem[{Veale and Hao(2008)}]{veale-2009}
Tony Veale and Yanfen Hao. 2008.
\newblock \href {https://www.aclweb.org/anthology/C08-1119} {A fluid knowledge
  representation for understanding and generating creative metaphors}.
\newblock In \emph{Proceedings of the 22nd International Conference on
  Computational Linguistics}, pages 945--952, Manchester, UK. COLING 2008
  Organizing Committee.

\bibitem[{Veale et~al.(2016)Veale, Shutova, and Klebanov}]{veale-2016-2}
Tony Veale, Ekaterina Shutova, and Beata~Beigman Klebanov. 2016.
\newblock \href {https://doi.org/10.2200/S00694ED1V01Y201601HLT031} {Metaphor:
  A computational perspective}.
\newblock \emph{Synthesis Lectures on Human Language Technologies},
  9(1):1--160.

\bibitem[{Wallington et~al.(2011)Wallington, Agerri, Barnden, Lee, and
  Rumbell}]{wallington-2011}
Alan Wallington, Rodrigo Agerri, John Barnden, Mark Lee, and Tim Rumbell. 2011.
\newblock \href {https://doi.org/10.1007/978-94-007-1757-2_5} {Affect transfer
  by metaphor for an intelligent conversational agent}.
\newblock In \emph{Affective computing and sentiment analysis. Emotion,
  metaphor and terminology}, volume~45, pages 53--66.

\bibitem[{Yu and Wan(2019)}]{yu-2019}
Zhiwei Yu and Xiaojun Wan. 2019.
\newblock \href {https://doi.org/10.18653/v1/N19-1092} {How to avoid sentences
  spelling boring? {T}owards a neural approach to unsupervised metaphor
  generation}.
\newblock In \emph{Proceedings of the 2019 Conference of the North {A}merican
  Chapter of the Association for Computational Linguistics: Human Language
  Technologies}, pages 861--871, Minneapolis, Minnesota. Association for
  Computational Linguistics.

\bibitem[{Zhang(2008)}]{zhang-2008}
Li~Zhang. 2008.
\newblock \href {https://doi.org/10.1145/1501750.1501773} {Metaphorical affect
  sensing in an intelligent conversational agent}.
\newblock In \emph{Proceedings of the Fifth International Conference on
  Advances in Computer Entertainment Technology}, pages 100--106, Yokohama,
  Japan.

\end{thebibliography}
\bibliographystyle{tacl2020}

\end{document}